\documentclass[10pt,twocolumn,letterpaper]{article}

\usepackage{cvpr}              

\usepackage[dvipsnames]{xcolor}

\definecolor{cvprblue}{rgb}{0.21,0.49,0.74}
\usepackage[pagebackref,breaklinks,colorlinks,citecolor=cvprblue]{hyperref}
\usepackage{graphicx}
\usepackage{capt-of}
\usepackage{comment}
\usepackage{amsmath,amssymb}
\usepackage{color}
\usepackage{url}
\usepackage{hyperref}
\usepackage{cite}
\usepackage{amsmath,amssymb,amsfonts}
\usepackage{algorithmic}
\usepackage{graphicx}
\usepackage{textcomp}
\usepackage{xcolor}
\usepackage{times}
\usepackage{epsfig}
\usepackage{graphicx}
\usepackage{amsmath}
\usepackage{amssymb}
\usepackage{graphicx}
\usepackage{amsmath,makecell}
\usepackage{amssymb}
\usepackage{booktabs}
\usepackage{multirow}
\usepackage{subcaption}
\usepackage{array}
\usepackage{stackengine}
\usepackage{amssymb}
\usepackage{booktabs}
\usepackage{rotating}
\usepackage{utfsym}
\newcolumntype{C}[1]{>{\centering\let\newline\\\arraybackslash}m{#1}}
\newcolumntype{L}[1]{>{\raggedright\let\newline\\\arraybackslash}m{#1}}
\usepackage[capitalize]{cleveref}
\crefname{section}{Sec.}{Secs.}
\Crefname{section}{Section}{Sections}
\Crefname{table}{Table}{Tables}
\crefname{table}{Tab.}{Tabs.}
\newcommand\xnew{0.24\textwidth}
\newcommand\ynew{5.6em}
\newcommand\x{0.119\textwidth}
\newcommand\y{8.5em}

\newcommand\xx{0.16\textwidth}

\def\paperID{165} 
\def\confName{3DV\xspace}
\def\confYear{2024\xspace}

\title{GenLayNeRF: Generalizable Layered Representations with 3D Model Alignment for Human View Synthesis}

\author{Youssef Abdelkareem\\
University of Waterloo\\
{\tt\small yafathi@uwaterloo.ca}
\and
Shady Shehata\\
MBZUAI\\
{\tt\small shady.shehata@mbzuai.ac.ae}
\and
Fakhri Karray\\
University of Waterloo, MBZUAI\\
{\tt\small karray@uwaterloo.ca}
}

\begin{document}
\maketitle

\begin{abstract}
Novel view synthesis (NVS) of multi-human scenes imposes challenges due to the complex inter-human occlusions. Layered representations handle the complexities by dividing the scene into multi-layered radiance fields, however, they are mainly constrained to per-scene optimization making them inefficient.
Generalizable human view synthesis methods combine the pre-fitted 3D human meshes with image features to reach generalization, yet they are mainly designed to operate on single-human scenes. Another drawback is the reliance on multi-step optimization techniques for parametric pre-fitting of the 3D body models that suffer from misalignment with the images in sparse view settings causing hallucinations in synthesized views. In this work, we propose, GenLayNeRF, a generalizable layered scene representation for free-viewpoint rendering of multiple human subjects which requires no per-scene optimization and very sparse views as input. We divide the scene into multi-human layers anchored by the 3D body meshes. We then ensure pixel-level alignment of the body models with the input views through a novel end-to-end trainable module that carries out iterative parametric correction coupled with multi-view feature fusion to produce aligned 3D models. For NVS, we extract point-wise image-aligned and human-anchored features which are correlated and fused using self-attention and cross-attention modules. We augment low-level RGB values into the features with an attention-based RGB fusion module. To evaluate our approach, we construct two multi-human view synthesis datasets; DeepMultiSyn and ZJU-MultiHuman. The results indicate that our proposed approach outperforms generalizable and non-human per-scene NeRF methods while performing at par with layered per-scene methods without test time optimization.

\end{abstract}

\section{Introduction}


Novel view synthesis (NVS) of scenes with human subjects has numerous applications in telepresence, virtual reality, etc. The extensions \cite{Gao2020PortraitNR,Li2021Neural3V,Xian2021SpacetimeNI,Pumarola20arxiv_D_NeRF} of the well-known NeRF \cite{Mildenhall2020NeRFRS} architecture achieved competitive synthesis results using sparse views, yet suffered with human subjects due to their complex motions. NeuralBody \cite{Peng2021NeuralBI} anchored NeRF with pre-fitted 3D human models to regularize the training producing more photo-realistic output. A main constraint was the inefficient per-scene optimization requirement. 
Recently, state-of-the-art human-based synthesis methods \cite{Kwon2021NeuralHP,Zhao2021HumanNeRFGN,Mihajlovi2022KeypointNeRFGI,Cheng2022GeneralizableNP} merged the concepts of the human model anchors and the image features to generalize to unseen poses and human identities. However, they were only designed to operate on scenes with single human subjects. Multi-human scenes introduce additional challenges due to how humans occlude each other and the complexity of their close interactions. Layered scene representations \cite{Zhang2021EditableFV} are a possible solution to operate in the complex multi-person setting. 
Shuai et al. \cite{Shuai2022NovelVS} utilized a layered architecture by representing the human entities using NeuralBody \cite{Peng2021NeuralBI} and weakly supervising the human instance segmentation. Nevertheless, the method suffers from the per-scene optimization problem which hinders its applicability to wider real-world domains. Another issue with existing Human NVS methods \cite{Kwon2021NeuralHP,Shuai2022NovelVS,Zhao2021HumanNeRFGN} is the reliance on multi-step optimization methods \cite{Bogo2016KeepIS,Zhang2021LightweightMT,easymocap} for the estimation of pre-fitted 3D body models. Such methods hinder the ability of end-to-end learning and suffer from error accumulation throughout the fitting steps which lead to inaccurate parameter fitting and misaligned body models and consequently hurts the synthesis quality of the novel views.

In this paper, we
propose generalizable layered neural radiance fields to achieve free-viewpoint rendering of multi-human subjects, while requiring no test-time optimization for novel subjects or poses.  We fuse the concepts of implicit feature aggregation and layered scene representations to synthesize novel views of complex human interactions from very sparse input streams. Specifically, we divide the scene into a set of human layers anchored by the 3D human body meshes. We then introduce a novel end-to-end trainable human-image alignment module that utilizes an iterative feedback loop \cite{pymaf} to correct parametric errors in the pre-fitted human models and produces pixel-aligned human layers for better synthesis quality.  For view synthesis, we extract a set of point-wise image-aligned and human-anchored features for all views and effectively aggregate them using self-attention and cross-attention modules. We also include an RGB fusion module that embeds the fused features with low-level pixel information from the images for retaining high-frequency details. 



Our main contributions are summarized as follows:
\begin{itemize}
    \item We propose a generalizable layered representation with a novel combination of three attention-based feature fusion modules for free-viewpoint rendering of multi-human scenes from sparse input views while operating on novel human subjects and poses.
   
    \item We present a novel human-image alignment module that corrects misalignment errors in the pre-fitted human models through an end-to-end trainable iterative feedback loop coupled with multi-view self-attention feature fusion.
    
    \item We surpass state-of-the-art generalizable and non-human per-scene NeRF methods while performing at par with the multi-human per-scene methods without requiring long per-scene training procedures.

\end{itemize}

\section{Related Work}

\subsection{Neural View Synthesis}
Recent progress has been made in utilizing neural networks along with differentiable rendering for novel view synthesis \cite{Sitzmann2019DeepVoxelsLP,Yan2016PerspectiveTN,Aliev2020NeuralPG,Wu2020MultiViewNH,Flynn2019DeepViewVS,Li2021MINETC,Thies2019DeferredNR}. 
NeRF \cite{Mildenhall2020NeRFRS} encapsulated the full continuous 5D radiance field of scenes inside a Multi-Layer Perceptron (MLP). 
They achieved photo-realistic results but failed to work on highly deformable scenes with non-static subjects. Deformable NeRF methods \cite{Park20arxiv_nerfies,Pumarola20arxiv_D_NeRF} modeled the dynamic subjects by training a deformation network that transforms 3D points to a canonical space before querying the MLP. Yet, they show poor synthesis quality for human subjects with complex deformations. NeuralBody \cite{Peng2021NeuralBI} anchored NeRF with a deformable human model \cite{SMPL:2015} to provide a prior over the human body shape and correctly render self-occluded regions. However, they lacked generalization capabilities for novel scenes.
Per-scene optimization NeRF methods \cite{Peng2021NeuralBI,Mildenhall2020NeRFRS,Pumarola20arxiv_D_NeRF,Shuai2022NovelVS,Peng2021NeuralBI} need to be trained from scratch on each scene  which is often impractical due to the large time and computational costs. Generalizable NeRF methods \cite{Trevithick2020GRFLA,Yu2021pixelNeRFNR,Wang2021IBRNetLM} offer a solution by conditioning NeRF on pixel-aligned features generated from the input images which enhanced the results for unseen scenes with sparse input views. 
Recently, NHP \cite{Kwon2021NeuralHP} combined the 3D human mesh with image features to accurately represent complex body dynamics and generalize to novel human subjects and poses. 
HumanNeRF \cite{Zhao2021HumanNeRFGN} enhanced the quality through efficient fine-tuning procedures and  neural appearance blending techniques. However, the blending module operates on pre-scanned synthetic data with accurate depth maps and cannot be extended to real-world data. One limitation of state-of-the-art generalizable human methods \cite{Kwon2021NeuralHP,Zhao2021HumanNeRFGN,Cheng2022GeneralizableNP} lies in the inability to be extended to multi-human scenes which are challenging due to the inter-human occlusions and interactions.

Layered scene representations \cite{Lu2020LayeredNR} were proposed to handle complex scenes with multiple human subjects.
ST-NeRF
\cite{Zhang2021EditableFV} modeled each human layer using a deformable model similar to D-NeRF \cite{Pumarola20arxiv_D_NeRF} to achieve editable free-viewpoint rendering. Recently, Shuai et al. \cite{Shuai2022NovelVS} extended ST-NeRF by modeling the human subjects using NeuralBody \cite{Peng2021NeuralBI} and predicted human segmentation masks as part of the network training. 
The restriction of both methods is requiring per-scene training procedures for learning, yielding them inefficient to use. We tackle the existing research gap by proposing a generalizable layered scene representation for synthesizing novel views of multi-human subjects through a combination of image features and  layered neural radiance fields. We achieve free-viewpoint rendering for scenes with an arbitrary number of humans from very sparse input views, while generalizing to novel subjects and poses at test time without extra optimization. 

\label{related:nvc}
\begin{figure*}[t]
\centering
\includegraphics[width=0.99\textwidth]{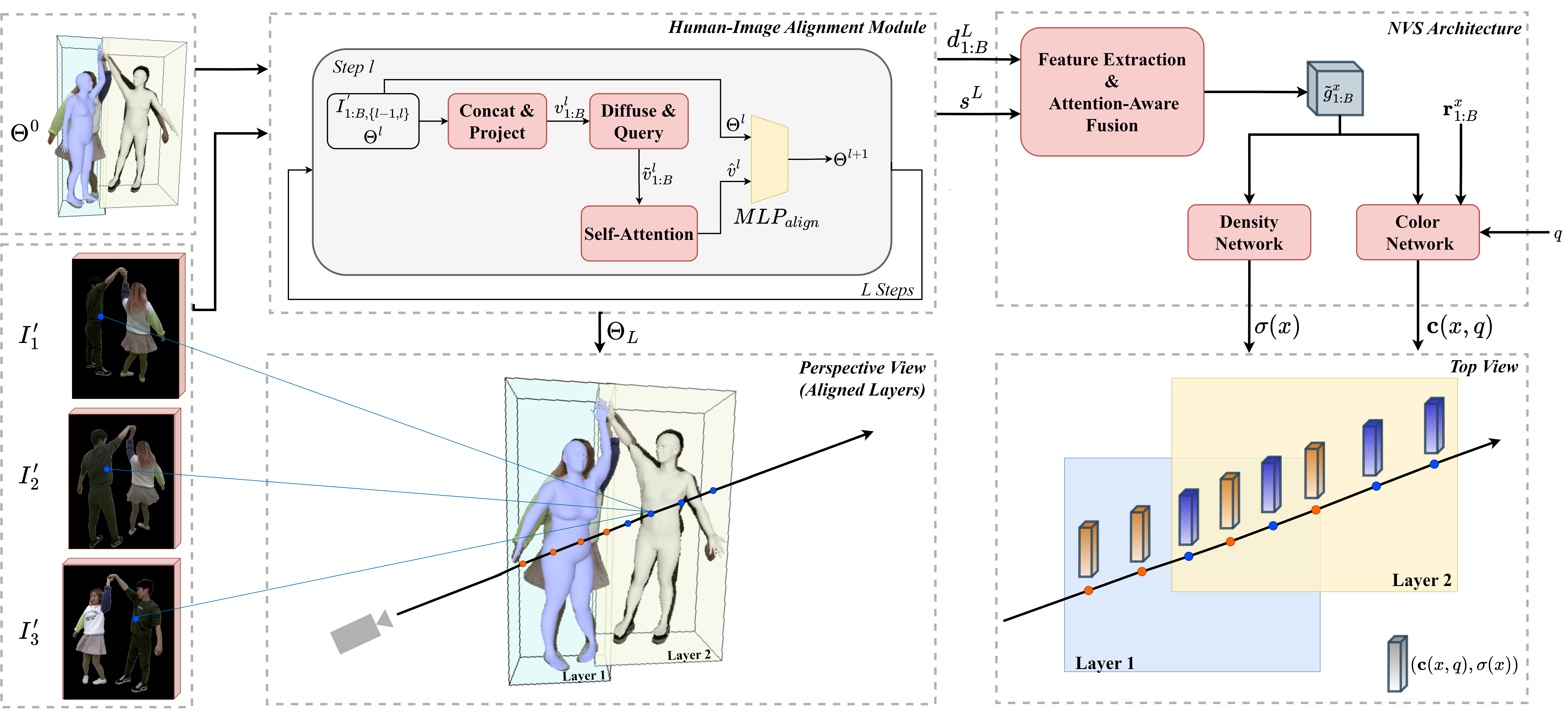}
\caption{Overview of the GenLayNeRF approach. 
We consolidate a layered scene representation where each human subject is modeled using the SMPL model. 
Regarding our alignment module, at a step $l$, low and high-resolution feature planes $I'_{1:B,\{l-1,l\}}$ are concatenated  and the SMPL vertices are projected on them to produce feature-embedded vertices $v^{l}_{1:B}$ (Concat \& Project). We then diffuse the vertices to continuous spaces and query them at downsampled vertex locations to generate multi-view human features $\Tilde{v}^l_{1:B}$ (Diffuse \& Query), which are fused using self-attention and passed along with parameters $\Theta^l$ to predict the adjusted parameters $\Theta^{l+1}$.
In our NVS architecture, we project rays through the aligned scene layers and sample per-layer 3D points within the intersections areas with the layers (shown in the top view). Point-wise features are extracted and fused to output the final fused features $\Tilde{g}^x_{1:B}$, which are passed to the density network to predict the volume density $\sigma(x)$, whereas the color network uses the raw RGB values $\mathbf{r}^x_{1:B}$ and $q$ to predict the color $\mathbf{c}(x,q)$.}
\label{fig:arch}
\end{figure*}
\subsection{Human Mesh Recovery}
\label{related:hpc}
Mesh Recovery of human subjects has grabbed significant research attention due to its adoption in 3D geometry reconstruction and novel view synthesis. One direction of approaches solves the task through a multi-step optimization process which fits the parametric human models (i.e. SMPL \cite{SMPL:2015}) based on 2D observations such as keypoints or silhouettes \cite{Guan2009EstimatingHS,Sigal2007CombinedDA}. Bogo et al. \cite{Bogo2016KeepIS} utilized 2D joint predictions from monocular input to guide the SMPL fitting process for single-human scenes. Zhang et al. \cite{Zhang2021LightweightMT} tackled a more challenging multi-person setting by leveraging triangulated 3D keypoints and a two-step parametric fitting process for enhanced results. The main issues with multi-step methods are breaking the end-to-end learning and the error accumulation throughout the steps, especially in sparse-view datasets. Specifically, 2D keypoints predictions could suffer from inaccurate joints in certain views which hurts the triangulation process leading to low-quality 3D keypoint predictions. The parametric model fitting is subject to errors due to the abundance of hyperparameters \cite{Zhang2021LightweightMT} that require meticulous finetuning and the accumulated errors from the previous steps. On the other hand, regression-based approaches aim for better human-image alignment by directly regressing the body models from input images \cite{pymaf,Zanfir2020WeaklyS3,Kundu2020AppearanceCD,Lin2021MeshG,Lin2020EndtoEndHP,Kocabas2021PAREPA}. PyMAF \cite{pymaf} introduced a feedback loop with multi-scale contexts to correct parametric deviations for producing highly aligned meshes from monocular input images for single-humans.

Existing Human NVS approaches \cite{Kwon2021NeuralHP,Peng2021NeuralBI,Zhao2021HumanNeRFGN,Mihajlovi2022KeypointNeRFGI,Cheng2022GeneralizableNP} utilize pre-fitted 3D observations computed using multi-step optimization approaches \cite{Zhang2021LightweightMT,easymocap}. However, in sparse-view settings, the pre-fitted predictions suffer from misalignment errors that consequently hurt the quality of the synthesized views. Mihajlovi et al. \cite{Mihajlovi2022KeypointNeRFGI} utilized 3D keypoints instead of body models to avoid parametric fitting errors. L-NeRF \cite{Shuai2022NovelVS} introduced a time-synchronization step that accounts for the multi-view image de-synchronization by producing a per-view body model using predicted time offsets. However, they do not account for parametric errors occurring in the multi-step fitting process. In this work, we propose a novel regression-based human-image alignment module that ensures the correction of parametric errors leading to aligned body models with multi-view input. 

\section{Methodology}
\subsection{Problem Definition}
Given a synchronized set $\Omega$ of frames $I$ taken from $B$ sparse input viewpoints of a scene with $N$ arbitrary number of humans, such that $\Omega = {\{I_{1},..,I_{B}\}}$, our target is to synthesize a novel view frame ${\{I_{q}\}}$ of the scene from a query viewing direction $q$. Each input viewpoint $b$ is represented by the corresponding camera intrinsics $K$, and camera rotation $R$ and translation $t$, where $b= {\{K_b,[R_b|t_b]\}}$. The $N$ pre-fitted 3D human body meshes are given for each input frame. Each human $h$ is represented using the SMPL \cite{SMPL:2015} model  which is a deformable skinned model defined in terms of pose and shape parameters $\Theta_h^0$ while also being vertex-based where each model $s_h$ consists of 6,480 vertices, such that $s_h \in \mathbf{R}^{6,480 \times 3}$. For an input view image $I_b \in \mathbf{R}^{H \times W \times 3}$ with height $H$ and width $W$, we extract a multi-scale feature pyramid $I^{'}_{b,\{0:T-1\}}$ with $T$ levels  using a ResNet34 \cite{He2016DeepRL} backbone network $f$, pre-trained on ImageNet, such that $ I^{'}_{b,\{0:T-1\}} = f(I_b)$. The operation is carried out for all input views $b$ in $\{1,.., B\}$.
A full overview of the proposed architecture is shown in \cref{fig:arch}
.

\subsection{Human-Image Alignment Module}
Pre-fitted human body models can suffer from misalignment with the input images due to error accumulation throughout the multi-step fitting process \cite{Zhang2021LightweightMT}, especially in sparse view settings, which causes hallucinations in synthesized views. 
We propose an alignment module that is end-to-end trainable with our NVS architecture and carries out iterative parametric correction with closed feedback \cite{pymaf} to ensure a better alignment of the SMPL models with the multi-view input images. The module takes the pre-fitted SMPL parameters $\Theta_h^{0}$ as input and returns the aligned and adjusted parameters $\Theta_h^{L}$. Specifically, we employ an iterative process with $L$ steps, such that, for a step $l>0$, low-resolution features $I^{'}_{b,l-1}$ from level $l-1$ for view $b$ are upsampled using deconvolution \cite{Noh2015LearningDN} and concatenated with high-resolution feature plane $I^{'}_{b,l}$ at level $l$  resulting in a contextualized and localized feature plane $I^{''}_{b,l}$. Human vertices $s^{l}_h$ are embedded with image features by projection on the multi-view feature map, such that, $v_{h,b}^{l} = {I_{b,l}^{''}}[K_b((R_b s^{l}_h)+t_b)]$. $v_{h,b}^{l} \in \mathbf{R}^{6,480 \times C_1}$ represents the features of the vertices projected on feature map $I_{b,l}^{''}$ for human $h$. The preceding part corresponds to "Concat \& Project" in \cref{fig:arch}.

Our target is to retrieve a compact and continuous per-human feature representation to be used for parameter adjustment. For that reason, the sparse human vertices $v_{h,b}^{l}$ need to be diffused into a continuous space that can be queried at any location. We incorporate the SparseConvNet \cite{Graham20183DSS,Peng2021NeuralBI} architecture which utilizes 3D sparse convolution to diffuse the vertex features into different nearby continuous spaces for every human and view. The diffused vertices are denoted as $d_{h,b}^{l}$. To obtain the per-human features, we downsample the vertices $s^{l}_h$, such that $\Tilde{s}^{l}_h \in \mathbf{R}^{431 \times 3}$, and query the diffused vertex spaces at the downsampled locations to obtain the multi-view per-human vertex features which are then processed and flattened to obtain a compact version denoted as $\Tilde{v}_{h,b}^{l} \in \mathbf{R}^{1 \times C_2}$. The preceding part corresponds to "Diffuse \& Query" in \cref{fig:arch}. Afterward, we effectively correlate the multi-view human features using a self-attention module, such that,
\begin{equation}
    \begin{gathered}
    mv_h^{l} = soft(\frac{1}{\sqrt{d_{k_1}}}query(\Tilde{v}_{h,{1:B}}^{})\ .\ key(\Tilde{v}_{h,{1:B}}^{l})^T) , \\
    \hat{v}_{h,{1:B}}^{l} = mv_h^{l}\ .\  val_1(\Tilde{v}_{h,{1:B}}^{l}) + val_2(\Tilde{v}_{h,{1:B}}^{l})
    ,\\ \ mv_h^{l} \in \mathbf{R}^{B \times B},\  \Tilde{v}_{h,{1:B}}^{l} \in \mathbf{R}^{B \times C_2},
    \end{gathered}
\label{eq:self}
\end{equation}

\noindent where $key$, $query$, and $(val_{1},val_{2})$ represent the key, query, and value embeddings of the corresponding argument features respectively, and $d_{k_1}$ denotes the dimensionality of the key embedding. $soft$ denotes the softmax operation. We carry out view-wise averaging for multi-view fusion on the view-aware human features such that, $\hat{v}_{h}^{l} = \frac{1}{B} \sum_{b}{}{\hat{v}_{h,b}^{l}}$. Lastly, the fused per-human features are concatenated ($\oplus$) with the current SMPL parameters and passed to a correction MLP that predicts parameter alignment offsets $\Delta\Theta^{l}_{h}$ which are added to the current parameters, such that,
\begin{equation}
    \begin{gathered}
    \Delta\Theta^{l}_{h} = MLP_{align}([\hat{v}_{h}^{l} \oplus  \Theta^{l}_{h}]), \\
    \Theta^{l+1}_{h} = \Theta^{l}_{h} + \Delta\Theta^{l}_{h},
    \end{gathered}
\end{equation}

\noindent The updated parameters $\Theta^{l+1}_{h}$ are used to retrieve the adjusted SMPL vertices $s^{l+1}_h$ and are passed to the next step $l+1$. After $L$ steps, the aligned SMPL parameters $\Theta^{L}_{h}$, vertices  $s^{L}_h$, and diffused spaces $d_{h,{1:B}}^{L}$ are passed to our layered NVS architecture.


\subsection{Layered Scene Representation}
Scenes with multiple humans suffer from inter-human occlusions that become evident when subjects closely interact together. A practical solution to handle complex multi-human scenarios is dividing the scene into distinct layers where each layer models an entity using a neural radiance field \cite{Zhang2021EditableFV,Lu2020LayeredNR}. Entities can be humans, objects, or background. Our proposed approach focuses mainly on human layers represented using the SMPL \cite{SMPL:2015} model which is responsible for preserving the local geometry and appearance of humans making it possible to model their complex deformations and occluded areas.

Our aim is to render the full novel view image $I_q$ from a query viewpoint $q$. To achieve that, we first use the camera-to-world projection matrix, defined as $\mathbf{P}^{-1} = [R_q|t_q]^{-1}K_q^{-1}$, to march 3D rays across the multi-layered scene. In practice, we have a ray for each pixel $p$ in the final image, where the ray origin $r_0 \in \mathbf{R}^3$ is the camera center and the ray direction is given as $i = \frac{\mathbf{P}^{-1}p-r_0}{||\mathbf{P}^{-1}p-r_0||}$. 3D points $x$ are sampled across the rays at specific depth values $z$, where $x = r(z) =  r_0 + z i$. Since we have several human layers in the scene, we determine the intersection areas of the rays with the humans using the 3D bounding box around each layer defined by the minimum and maximum vertex points of the aligned SMPL meshes $s^{L}_{1:N}$.
We then sample depth values within the $n_p$ intersecting areas only such that $z \in [[z_{near_1},z_{far_1}],..,[z_{near_{n_p}},z_{far_{n_p}}]]$. This guarantees that the sampled points lie within the areas of the relevant human subjects as clear in the top view shown in \cref{fig:arch}. 

\subsection{Feature Extraction and Attention-Aware Fusion}

In our proposed approach, we extract multi-view image features for each query point $x$ and effectively merge them using attention-based fusion modules to derive the needed spatially-aligned feature vectors. 
This enables us to extrapolate to novel human subjects and poses 
by learning implicit correlations between the independent human layers.

\subsubsection{Image-aligned And Human-anchored Features}

Image-aligned point-wise features are extracted by projecting the point $x$ on all the feature maps $I^{''}_{b,L}$ to collect the corresponding image-aligned features for each view $b$ denoted as $p^x_b$. In addition, human-anchored features are beneficial for maintaining the complex geometric structure of the human body by anchoring the network on the available SMPL body priors. Existing layered scene representations \cite{Shuai2022NovelVS} follow the approach of NeuralBody \cite{Peng2021NeuralBI} by encoding the vertices of human layers using learnable embeddings that are unique to each layer in each training scene. In our approach, we utilize the vertices $v_{{1:N},{1:B}}^{L}$  embedded with image features from the alignment module to enable a generalizable approach conditioned on the input images. The radiance field predictor is queried using continuous 3D sampled points. For that reason, we utilize the diffused vertex spaces $d_{h,{1:B}}^{L}$ for each human $h$ and transform $x$ to the SMPL coordinate space of its corresponding human layer. Trilinear interpolation is then utilized to retrieve the corresponding human-anchored features $g^x_{b}$ from the diffused spaces of each view $b$.

\subsubsection{Attention-Aware Feature Fusion}
\label{att_aware}

To fuse the point-wise feature representations $g^x_{1:B}$, $p^x_{1:B}$  for point $x$, one strategy is a basic averaging approach \cite{Saito2019PIFuPI,Saito2020PIFuHDMP}. This leads to smoother output and ineffective utilization of the information seen from distinct views. 
To learn effective cross-view correlations, we employ a self-attention module that attends between all the multi-view human-anchored features $g^x_{1:B}$ where each feature in one view is augmented with the extra features seen from the other views. Each view feature is first concatenated with its corresponding viewing direction $d'_b$. The formulation is the same as the one shown in \cref{eq:self}. The produced view-aware human-anchored features are denoted as $\hat{g}^{x}_{{1:B}}$.

We additionally make use of the rich spatial information in the image-aligned features by carrying out cross-attention from the view-aware human-anchored features to the image-aligned features. The similarity between the multi-view image features and the per-view vertex features is used to re-weigh the image features and embed them with the vertex features. The fused features $\Tilde{g}^{x}_{{1:B}}$ are calculated with a formulation similar to \cref{eq:self}. 
The detailed formulation of our cross-attention and self-attention modules are shown in the supplementary material.
Afterward, we carry out view-wise averaging, such that $\Tilde{g}^{x} = \frac{1}{B} \sum_{b}{}{\Tilde{g}^{x}_{b}}$, to generate the final fused feature representation for $x$. 


\subsection{Radiance Field Predictor}
\label{nerf_predictor}
\textbf{Color Network}.
To predict the color $\mathbf{c}$ of point $x$, we use the query viewing direction $
q$ to model the view-dependent effects \cite{Mildenhall2020NeRFRS}. In addition, we explicitly augment the high-level features with low-level pixel-wise information to leverage the high-frequency details in the images. This has been achieved with an RGB fusion module which concatenates the high-level features with the encoded raw RGB values $\mathbf{r}^x_{b}$ for each view $b$. RGB values from closer input views are assigned higher weights by cross-attending $q$ with the input viewing directions $d'_{1:B}$ such that,
\begin{equation}
    \begin{gathered}
    \Tilde{c}^x = MLP_{c_1}(\Tilde{g}^{x}_{1:B};\gamma(q);p^x_{1:B}), 
    \\
    \hat{c}^x_{1:B} = 
    \{[\Tilde{c}^x
    \oplus \gamma(\mathbf{r}^x_{1})],..., [\Tilde{c}^x
    \oplus \gamma(\mathbf{r}^x_{B})]\},\\
    rgb_{att}^{x} = soft(\frac{1}{\sqrt{d_{k_2}}}query(q)\ .\ key(d'_{1:B})^T) , \\
     \mathbf{c}(x,q) = MLP_{c_2}( rgb_{att}^{x}\ .\  val_1(\hat{c}^x_{1:B}))
    ,\\
    \ rgb_{att}^{x} \in \mathbf{R}^{1 \times B},
   \\
    \end{gathered}
\end{equation}

\textbf{Density Network}. We predict volume density $\sigma(x)$ for point $x$ using the fused feature $\Tilde{g}^x$, such that, $\sigma(x) = MLP_{\sigma}(\Tilde{g}^{x})$. 

$MLP_{\sigma}$,  $MLP_{c_1}$, and $MLP_{c_2}$ consist of fully connected layers described in the supplementary material. $\gamma:\mathbf{R}^{3} \rightarrow \mathbf{R}^{(6 \times l)+3 }$ denotes a positional encoding \cite{Mildenhall2020NeRFRS} with $2 \times l$ basis functions and $d_{k_2}$ is set to 16.

\subsection{Layered Volumteric Rendering and Loss Functions}

 Layered volumetric rendering is used to accumulate the predicted RGB and density for all points across human layers. The points in intersecting areas $n_p$ the layers are sorted based on their depth value $z$ before accumulation. The detailed formulation is shown in the supplementary material. Given a ground truth novel view image $I^{gt}_q$, all network weights are supervised using the L2 Norm  ($||.||$) photo-metric loss. In addition, we include two losses to explicitly supervise the training of our alignment module weights. Given a set of pseudo ground truth 2D keypoints $J^{gt}$, we derive the predicted 2D keypoints  $\Tilde{J}$ from the adjusted vertices $s^{L}$ following PyMAF \cite{pymaf} and minimize the keypoint difference weighted by the ground truth confidence of each body joint. We also include a regularization term on the SMPL parameters to avoid large parametric deviations. The final loss function for our network is written as,
\begin{equation}
    \mathbf{L} = \lambda_{ph} ||I^{gt}_q-I_q|| + \lambda_{kpts} ||J^{gt}-\Tilde{J}|| + \lambda_{reg} ||\Theta^{L}||,
\end{equation}

\section{Experiments}
In this section, we introduce the datasets, baselines, experimental results, and ablation studies. Details about our training procedure are in the supplementary material.

\begin{table}[t]
\caption{Comparison with generalizable and per-scene NeRF methods on the DeepMultiSyn and ZJU-MultiHuman Datasets. "G" and "S" denote generalizable and per-scene methods, respectively. "*" refers to human-based methods. PSNR and SSIM metric values are the greater the better. "ft" refers to finetuning.}
\begin{tabular}[t]{|c| l|c|c|c|c|}
\hline
& \multirow{2}{*}{Method} &  \multicolumn{2}{c|}{DeepMultiSyn} &\multicolumn{2}{c|}{ZJUMultiHuman} \\
\cline{3-6}
& & PSNR  & SSIM & PSNR  & SSIM  \\
\hline
\multicolumn{6}{|l|}{\textit{(a) Seen Models, Seen Poses}}
\\

\hline

    \multirow{4}{*}{\textit{S}} & NeRF & 15.49 & 0.497 & 16.42 & 0.525 \\
    & D-NeRF & 17.08 & 0.702 & 18.53 & 0.748  \\
      &  L-NeRF* & 24.04 & 0.858 & 25.10 & 0.903  \\
         \cline{2-6}
     
    &  \textbf{Ours$_{ft}$} & \textbf{25.05} & \textbf{0.889} &
     \textbf{25.21} & \textbf{0.916}  \\
     
     
     \hline
     \multirow{5}{*}{\textit{G}} & PixelNeRF & 14.81 & 0.534 & 19.74 & 0.629   \\
    & SRF & 20.39 & 0.724 & 17.87 & 0.657 \\
    & IBRNet & 19.45 & 0.741  & 20.03 & 0.766  \\
    & NHP* & 20.91 & 0.698 & 21.75 & 0.813  \\
     \cline{2-6}
     
    &  \textbf{Ours} & \textbf{24.01} & \textbf{0.859} &
     \textbf{25.02} & \textbf{0.901}  \\
     \hline
    \hline

\multicolumn{6}{|l|}{\textit{(b) Seen Models, Unseen Poses}}
\\

\hline
    \textit{S} & L-NeRF* & 22.12 & 0.825  & 23.02 & 0.871  \\
    %
     
     \hline
     \multirow{5}{*}{\textit{G}} & PixelNeRF & 14.14 & 0.520  & 16.88 & 0.560   \\
     & SRF & 18.07 & 0.663  & 17.93 & 0.680 \\
    & IBRNet & 18.01 & 0.710  & 19.84 & 0.772  \\
     & NHP* & 20.26 & 0.677 & 20.64 & 0.791  \\
     \cline{2-6}
     & \textbf{Ours} & \textbf{23.45} & \textbf{0.862} &
     \textbf{23.76} & \textbf{0.882}  \\
     \hline
    \hline

\multicolumn{6}{|l|}{\textit{(c) Unseen Models, Unseen Poses}}
\\
\hline

     \multirow{5}{*}{\textit{G}} & PixelNeRF & 13.12 & 0.457
     & \multicolumn{2}{c|}{\multirow{5}{*}{Not Applicable}}
     \\
    & SRF & 13.95 & 0.548 & \multicolumn{2}{c|}{}
     \\
    & IBRNet & 18.80 & 0.672 & \multicolumn{2}{c|}{} \\
    &   NHP* & 19.51 & 0.678 &   \multicolumn{2}{c|}{} \\
      \cline{2-4}
    &  \textbf{Ours} & \textbf{21.03} & \textbf{0.802} & \multicolumn{2}{c|}{}  \\
    \hline
\end{tabular}


\label{table:all_comp}

\end{table}
\subsection{Datasets}
The existence of readily-available open-source multi-human view synthesis datasets is limited. To solve this challenge, we construct two new datasets, ZJU-MultiHuman and DeepMultiSyn. Both datasets will be published to be used by multi-human view synthesis methods. We also include a subset of the single-human ZJU-MoCap dataset for diversity. Extra details on the datasets are included in the supplementary material.

\textbf{DeepMultiSyn.}
The DeepMultiSyn dataset is an adaptation of the 3D reconstruction dataset published by DeepMultiCap \cite{Zheng2021DeepMultiCapPC}. We take the raw real-world multi-view sequences and process them for novel view synthesis. There exist 3 video sequences of scenes containing 2 to 3 human subjects captured from 6 synchronized cameras. 
Following NeuralBody \cite{Peng2021NeuralBI}, we use EasyMoCap \cite{easymocap} to fit the SMPL human models for all the subjects in the available frames. Additionally, we predict the human segmentation masks following \cite{li2020self} to separate the humans from the background.
This dataset is considered challenging due to the existence of close interactions and complex human actions such as boxing, and dancing activities. 

\textbf{ZJU-MultiHuman.}
The ZJU-MultiHuman dataset consists of one video sequence with 600 frames taken from 8 uniformly distributed synchronized cameras. The video sequence was published online \cite{easymocap} with the calibration files. The captured scene contains 4 different human subjects. Similar to DeepMultiSyn, we predict the SMPL models and segmentation masks utilizing \cite{easymocap,li2020self}.  

\begin{figure*}[t]
\centering
\begin{tabular}[t]{c@{\hspace{1pt}}c@{\hspace{1pt}}c@{\hspace{1pt}}c}
{\includegraphics[width = \x,height=\y]{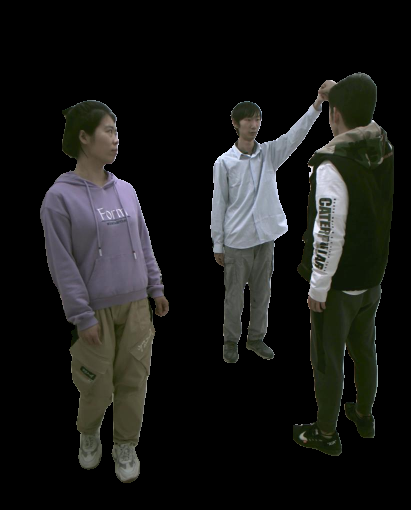}} &
{\includegraphics[width = \x,height=\y]{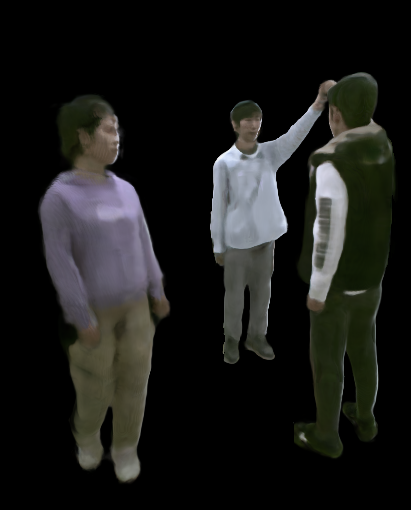}} &
{\includegraphics[width = \x,height=\y]{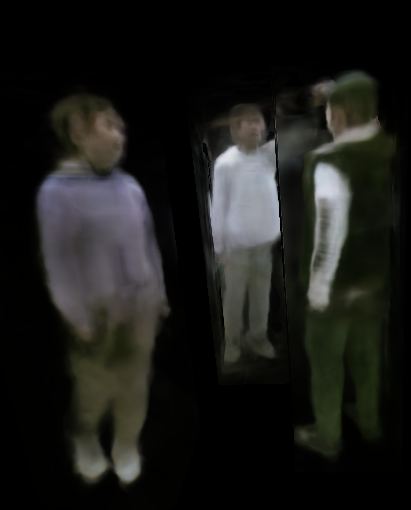}} &
{\includegraphics[width = \x,height=\y]{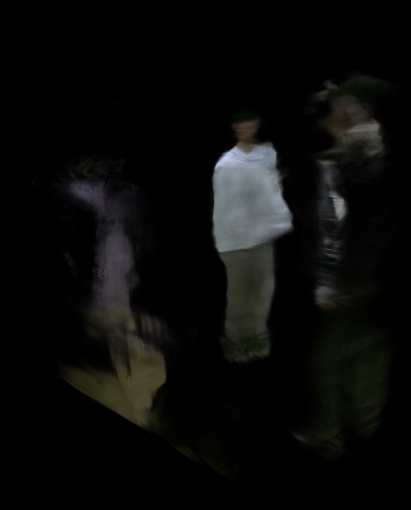}}\\
\stackunder[2pt]{\includegraphics[width = \x,height=\y]{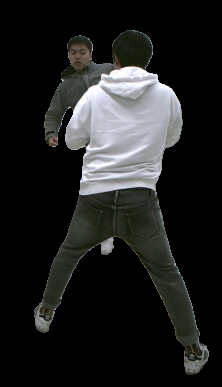}}{GT} 
&
\stackunder[2pt]{\includegraphics[width = \x,height=\y]{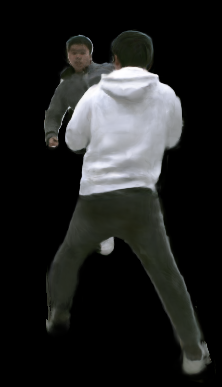}}{Ours} 
&
\stackunder[2pt]{\includegraphics[width = \x,height=\y]{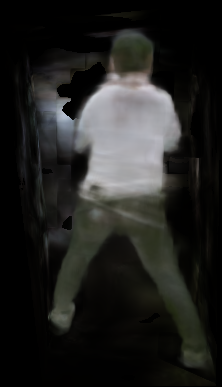}}{NHP}
&
\stackunder[2pt]{\includegraphics[width = \x,height=\y]{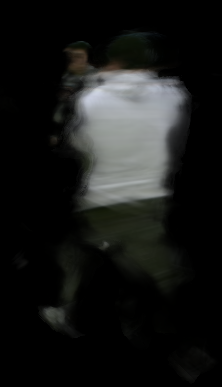}}{IBRNet}

\end{tabular}
\hspace{-1em}%
\begin{tabular}[t]{c@{\hspace{1pt}}c@{\hspace{1pt}}c@{\hspace{1pt}}c}
{\includegraphics[width = \x,height=\y]{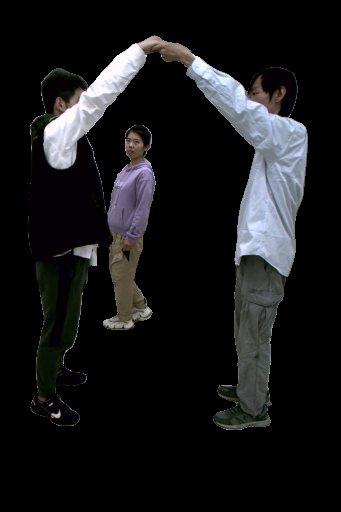}} &
{\includegraphics[width = \x,height=\y]{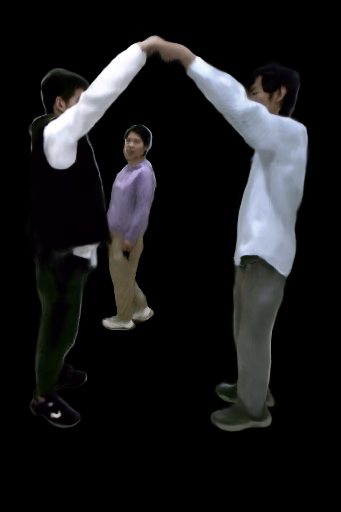}} &
{\includegraphics[width = \x,height=\y]{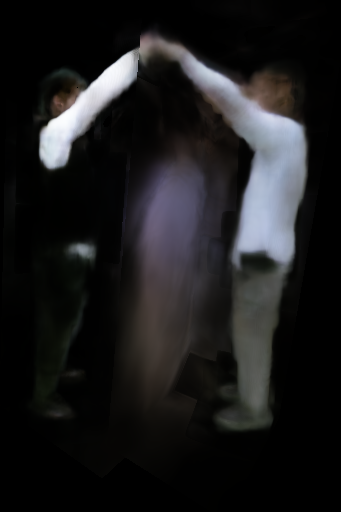}} &
{\includegraphics[width = \x,height=\y]{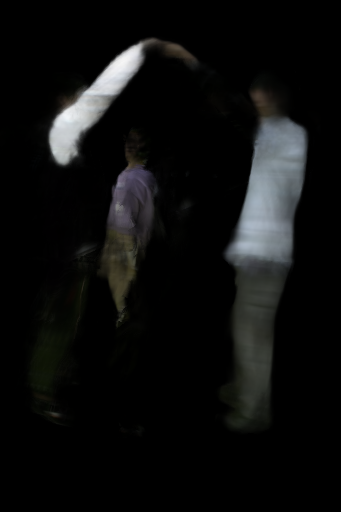}}\\
\stackunder[2pt]{\includegraphics[width = \x,height=\y]{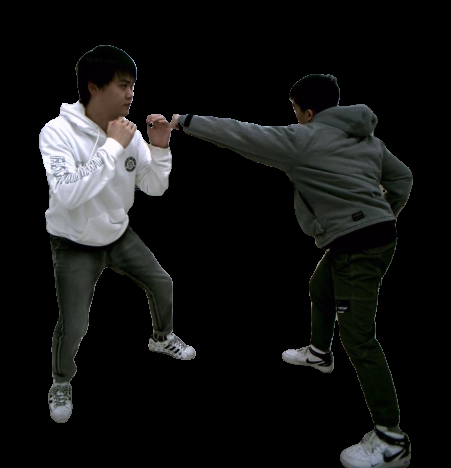}}{GT} &
\stackunder[2pt]{\includegraphics[width = \x,height=\y]{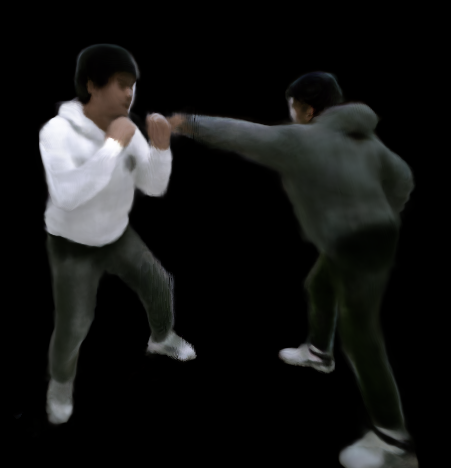}}{Ours} &
\stackunder[2pt]{\includegraphics[width = \x,height=\y]{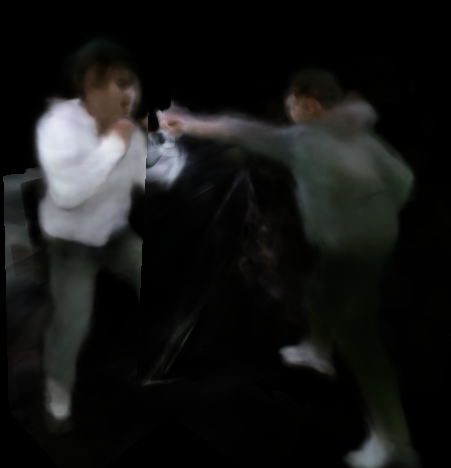}}{NHP}
&
\stackunder[2pt]{\includegraphics[width = \x,height=\y]{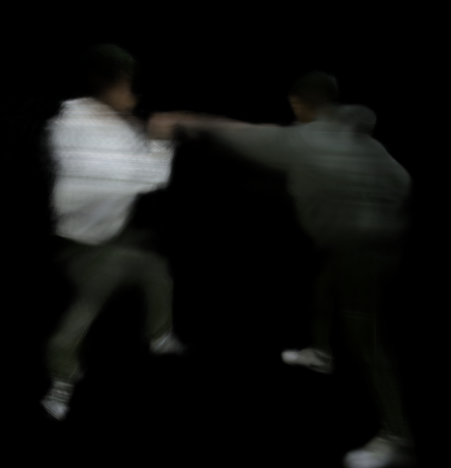}}{IBRNet} 
\end{tabular}

\caption{Comparison with generalizable methods on \textbf{seen models}/\textbf{unseen poses} [top row] and \textbf{unseen models}/\textbf{unseen poses} [bottom row] for the DeepMultiSyn Dataset.}
\label{fig:gen_main}

\end{figure*}

\begin{figure}[t]
\begin{tabular}[t]{c@{\hspace{1pt}}c@{\hspace{1pt}}c@{\hspace{1pt}}c@{\hspace{2pt}}}

\stackunder[2pt]{\includegraphics[width = \xx]{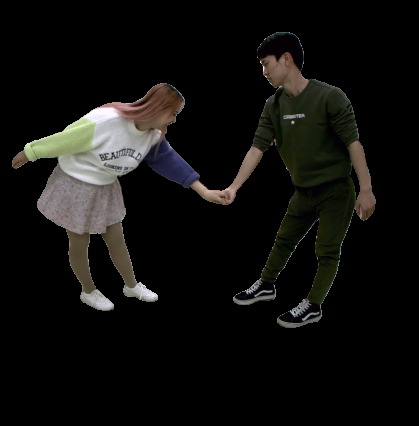}}{GT}
 & 
\stackunder[2pt]{\includegraphics[width = \xx]{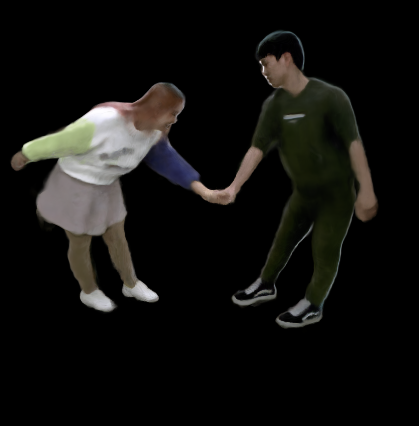}}{Ours} 
&
\stackunder[2pt]{\includegraphics[width = \xx]{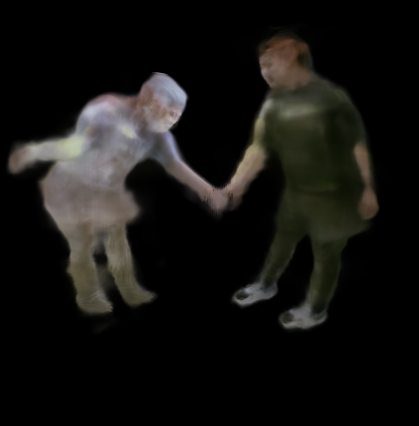}}{L-NeRF}
\end{tabular}

\caption{Comparison with a per-scene multi-human method \cite{Shuai2022NovelVS} on \textbf{seen models}/\textbf{unseen poses} on the DeepMultiSyn Dataset.}
\label{fig:gen_comp}
\end{figure}
\begin{figure}[t]
\begin{tabular}[t]{c@{\hspace{1pt}}c@{\hspace{1pt}}c@{\hspace{1pt}}c}
\stackunder[2pt]{\includegraphics[width = \xnew,height=\ynew]{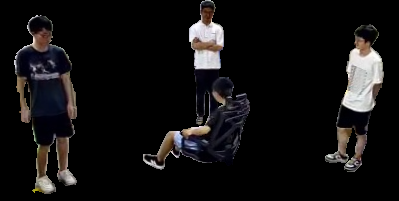}}{GT} &
\stackunder[2pt]{\includegraphics[width = \xnew,height=\ynew]{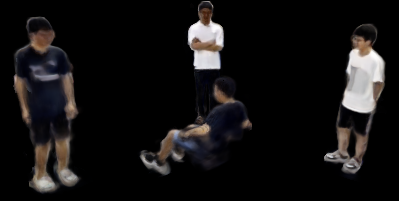}}{Ours} \\
\stackunder[2pt]{\includegraphics[width = \xnew,height=\ynew]{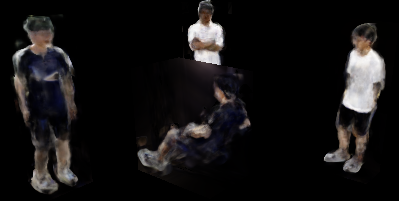}}{NHP}
&
\stackunder[2pt]{\includegraphics[width = \xnew,height=\ynew]{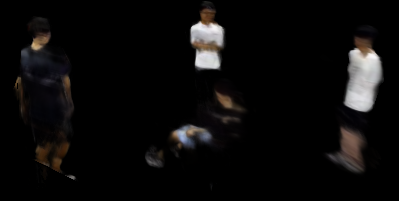}}{IBRNet} 
\end{tabular}

\caption{Comparison with generalizable methods on \textbf{seen models}/\textbf{unseen poses} for the ZJU-MultiHuman Dataset.}
\label{fig:gen_zju}
\end{figure}




\newcommand\xxxx{0.11\textwidth}
\newcommand\yyyy{8.5em}

\begin{figure}[t]
\centering
\begin{tabular}[t]{c@{\hspace{1pt}}c@{\hspace{1pt}}c@{\hspace{1pt}}c}

\stackunder[2pt]{\includegraphics[width = \xxxx,height=\yyyy]{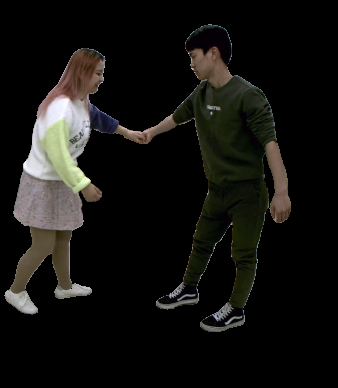}}{GT} &

\stackunder[2pt]{\includegraphics[width = \xxxx,height=\yyyy]{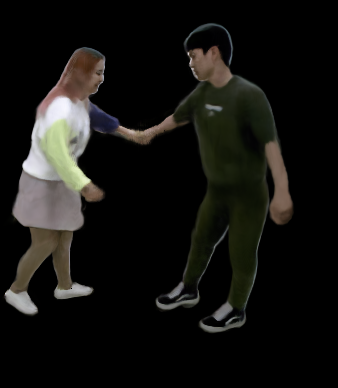}}{Ours}
&

\stackunder[2pt]{\includegraphics[width = \xxxx,height=\yyyy]{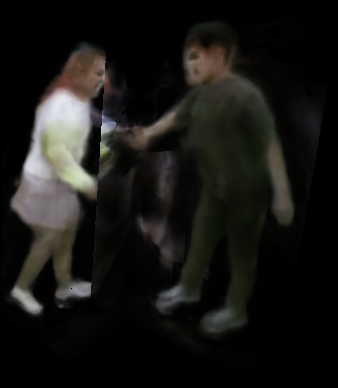}}{NHP}
&
\stackunder[2pt]{\includegraphics[width = \xxxx,height=\yyyy]{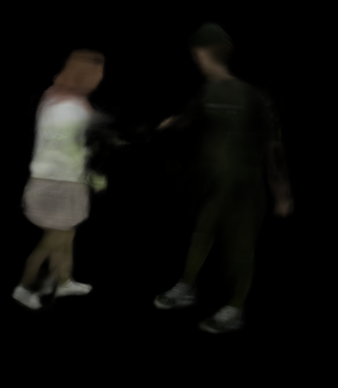}}{IBRNet}

\end{tabular}

\caption{Qualitative comparison on \textbf{unseen models/unseen poses} on the DeepMultiSyn dataset.}
\label{fig:gen_two_dance}

\end{figure}

\subsection{Baselines}
We compare our proposed approach with generalizable and per-scene NeRF methods.

\textbf{Comparison with generalizable NeRF methods.}
Generalizable human-based NeRF methods \cite{Kwon2021NeuralHP,Zhao2021HumanNeRFGN,Mihajlovi2022KeypointNeRFGI,Cheng2022GeneralizableNP} operate only on scenes with single humans. We choose to compare against NHP \cite{Kwon2021NeuralHP} after adjusting it to work on multi-human scenes by using the segmentation masks to render a separate image for each individual in the scene. We then superimpose the human images based on their depth to render the novel view image. Regarding non-human methods, PixelNeRF \cite{Yu2021pixelNeRFNR} is the first to condition NeRF on pixel-aligned features for generalization. IBRNet \cite{Wang2021IBRNetLM} and SRF \cite{SRF} additionally utilize image-based rendering and stereo correspondences, respectively, to achieve generalizable properties. 
All methods were trained on all human scenes simultaneously.

\textbf{Comparison with per-scene methods.}
 We evaluate our performance compared to the multi-human layered scene representation approach \cite{Shuai2022NovelVS}, denoted as L-NeRF.
We also compare against D-NeRF \cite{Pumarola20arxiv_D_NeRF} and the original NeRF \cite{Mildenhall2020NeRFRS} method. All of the mentioned approaches are trained on each scene separately with the same train-test splits.

\subsection{Experimental Results}
Our evaluation spans three generalization settings as follows:

\textbf{Seen Models, Seen Poses.} 
In this setting, we test on the same human subjects and poses that the model is trained on. \cref{table:all_comp}a indicates the results in terms of the per-scene and generalizable baselines. Regarding the generalizable approaches, our method exhibits the best overall performance on both datasets on all metrics. For the per-scene approaches,
our proposed method performs at par with the state-of-the-art per-scene baseline (L-NeRF), while effectively saving computational and time resources by taking 50 hours to converge on all the scenes simultaneously compared to 144 hours for per-scene training. After per-scene finetuning, our method surpasses L-NeRF on both datasets. Qualitative comparisons for the per-scene methods are included in the supplementary material.











\textbf{Pose Generalization.}
We additionally test all approaches on the same human subjects seen during training, but with novel poses. 
L-NeRF is a human-based method that generalizes to novel poses, therefore, we include it in the comparison.
On both datasets, \cref{table:all_comp}b shows that our approach outperforms all the generalizable NeRF methods on all metrics. 
L-NeRF lags behind our method on the DeepMultiSyn dataset due to the complex novel poses which validates the pose generalization ability of our method on challenging motions.  In \cref{fig:gen_main} and \cref{fig:gen_zju}, IBRNet fails to model the full body of the human subjects properly, while NHP fails to represent areas of occlusions where subjects highly overlap. However, our method successfully models the body shapes and can handle overlapping areas which validates the effectiveness of the layered scene representation in the generalizable multi-human setting. \cref{fig:gen_comp} shows how L-NeRF fails to properly render the appearance of subjects when presented with complex unseen poses. 


\textbf{Human Generalization.}
A challenging setting would be testing on human subjects and poses not seen during training. This was  done on the DeepMultiSyn dataset by leaving out two different scenes for testing. 
\cref{table:all_comp}c validates that our method has the best generalization capability as it outperforms all other methods by a large margin. The bottom row of \cref{fig:gen_main} and \cref{fig:gen_two_dance} show that our method better represents the main body features of the novel human subjects. IBRNet fails to render some body parts like the legs, while NHP suffers from more blur artifacts, especially in overlapping areas. In the supplementary material, we show that our method surpasses NHP by a large margin even on single humans in the ZJU-MoCap dataset for both pose and human generalization settings.

\newcommand\xxx{0.15\textwidth}
\newcommand\yyy{9em}

\begin{figure}[t]

\centering
\includegraphics[width=0.48\textwidth]{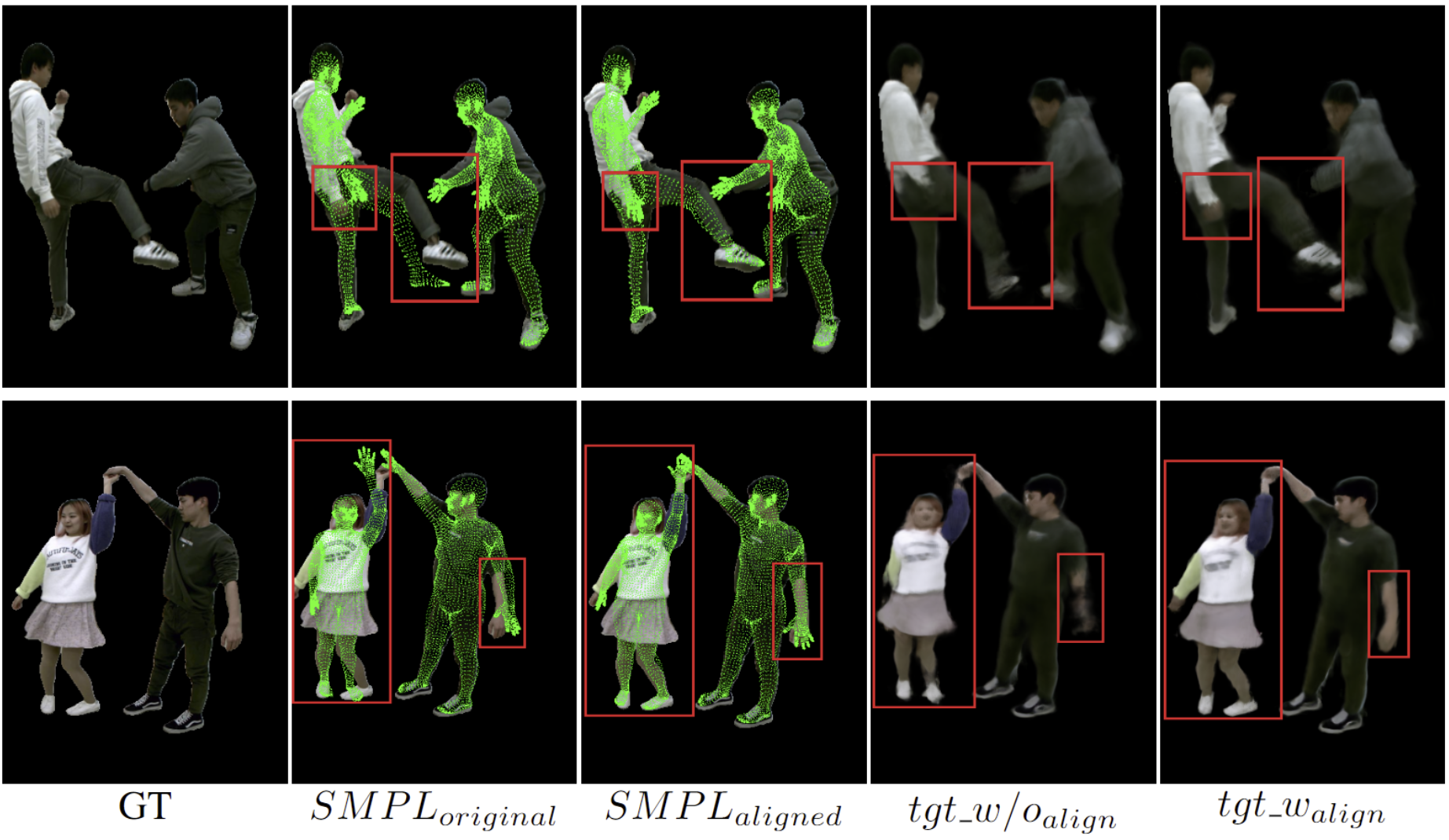}
\captionof{figure}{Visualization of the output of our human-image alignment module ($SMPL_{aligned}$) given the misaligned pre-fitted model ($SMPL_{original}$). "$tgt\_{w _{align}}$"  and "$tgt\_{w/o _{align}}$" denote the rendered image with/without our alignment module.}
\label{fig:alignment}
\end{figure}

\subsection{Ablation Studies}


\textbf{Effect of Human-Image Alignment.} We evaluate the impact of the proposed human-image alignment module on the synthesis quality. Quantitatively, \cref{table:ablation} shows the superior enhancement offered by the alignment module (align) on both metrics. In \cref{fig:alignment}, we demonstrate the large misalignment between the pre-fitted SMPL model and the image which caused severe hallucinations in the synthesized image (areas with red boxes). Our module successfully aligns the SMPL model with the images leading to higher-quality synthesis results. We include additional results of our module in the supplementary material. 

\textbf{Effect of Fusion modules.}
We assess the effect of different fusion modules on the synthesis results. From \cref{table:ablation}, the second row uses the cross-attention module (crs) in \cref{att_aware} and it shows a noticeable improvement over doing basic average pooling in the first row. This indicates the effectiveness of the correlation learned between the vertex and image features. The addition of the self-attention module (slf) in \cref{att_aware} in the third row led to the incorporation of multi-view aware features and achieved a slight enhancement on both metrics. The fourth row adds the raw RGB fusion module (rgb) in the Color Network presented in \cref{nerf_predictor}. It enhances the performance, especially on the SSIM metric, validating the importance of utilizing low-level information.


\textbf{Effect of Number of Views.}
We evaluate the performance of our proposed approach when given a different number of input views at test time. \cref{table:ablation} indicates that using 4 views leads to an enhancement in both metrics due to the extra information available. Decreasing the number of views gradually degrades the performance. However, using only one input view, our method outperforms all the generalizable NeRF methods in \cref{table:all_comp} that use 3 input views.


\section{Limitations \& Future Work}
Several enhancements to our proposed method could be investigated further. As our two proposed datasets were sufficient to show the generalization capability of our method, there is room for improvement by elevating the diversity in terms of the number of scenes, camera views, distinct humans, and complex actions. This would lead to better generalization capabilities on broader challenging scenarios. 
 Furthermore, our method suffers from blur artifacts representing human clothing details such as skirts as seen in \cref{fig:gen_comp}. One could experiment with integrating a deformation model \cite{Pumarola20arxiv_D_NeRF} to represent small deformations such as textured clothing. In addition, adjustments could be made to allow for human-image alignment for more complex body models such as SMPL-X \cite{SMPL-X:2019}. Lastly, a research direction could explore the optimization of the body model parameters from scratch with multi-view time synchronization taken into consideration.

\begin{table}[t]
\begin{tabular}[t]{|C{0.02\textwidth} C{0.02\textwidth} C{0.02\textwidth} C{0.03\textwidth} C{0.02\textwidth}|c c|}
\hline
      crs & slf & rgb & align & V. & PSNR $\uparrow$ & SSIM $\uparrow$  \\
    \hline
      & & & & 3 &20.92 & 0.7860 \\
    \hline
      \usym{1F5F8} & &  & & 3  & 21.45 & 0.8005 \\
    \hline
      \usym{1F5F8}& \usym{1F5F8}  & & & 3& 21.98 & 0.8093 \\
    \hline
      \usym{1F5F8} &\usym{1F5F8} & \usym{1F5F8}& & 3 &22.19 & 0.8361 \\
    \hline
      \usym{1F5F8} &\usym{1F5F8} & \usym{1F5F8}& \usym{1F5F8} & 3 &\textbf{23.45} & \textbf{0.8620} \\
    \hline
    \hline
    \usym{1F5F8} &\usym{1F5F8} & \usym{1F5F8}&\usym{1F5F8} & 1 &21.98 & 0.8091 \\
    \hline
    \usym{1F5F8} &\usym{1F5F8} & \usym{1F5F8}&\usym{1F5F8} & 2 &22.32 & 0.8379 \\
    \hline
    \usym{1F5F8} &\usym{1F5F8} & \usym{1F5F8}&\usym{1F5F8} & 4 & 23.72 & 0.8711 \\
    \hline
\end{tabular}
\label{table:ablation}
\centering
\captionof{table}{Ablation study results on \textbf{seen models} and \textbf{unseen poses} for the DeepMultiSyn dataset. "\# V." denotes the number of views.}
\end{table}

\section{Conclusion}

We introduce a generalizable layered scene representation for free-viewpoint rendering of multi-human scenes using very sparse input views while operating on unseen poses and subjects without test time optimization. We additionally present a novel end-to-end human-image alignment module that corrects parametric errors in the pre-fitted body models leading to pixel-level alignment of human layers with the input images. Regarding view synthesis, we divide the scene into a set of multi-human layers. We then generate point-wise image features and human-anchored features and utilize a combination of cross-attention and self-attention modules that effectively fuse the information seen from different viewpoints. In addition, we introduce an RGB fusion module to embed low-level pixel values into the color prediction for higher-quality results. We assess the efficacy of our approach on two newly proposed multi-human datasets. Experimental results show that our method outperforms state-of-the-art generalizable NeRF methods in different generalization settings and performs at par with layered per-scene methods without long per-scene optimization runs. We also validate the effectiveness of our alignment module by showing its significant enhancement on the synthesis quality. Our module could be integrated with existing SMPL-based synthesis methods to elevate the performance by improving the human-image alignment.

{
    \small
    \bibliographystyle{ieeenat_fullname}
    \bibliography{main}
}

\end{document}